\documentclass[10pt,twocolumn,letterpaper]{article}

\usepackage{cvpr}
\usepackage{times}
\usepackage{epsfig}
\usepackage{graphicx}
\usepackage{amsmath}
\usepackage{amssymb}
\usepackage{mathtools}
\usepackage{calligra}
\usepackage{color}
\usepackage{array}
\usepackage{hhline}
\usepackage{subfig}
\usepackage{multirow}
\usepackage{caption}
\usepackage{boxhandler}
\usepackage{tabularx}
\usepackage{adjustbox}
\usepackage[justification=centering]{caption}

\newcolumntype{P}[1]{>{\centering\arraybackslash}p{#1}}
\newcolumntype{M}[1]{>{\centering\arraybackslash}m{#1}}

\usepackage[pagebackref=true,breaklinks=true,letterpaper=true,colorlinks,bookmarks=false]{hyperref}

\cvprfinalcopy % *** Uncomment this line for the final submission

\def\cvprPaperID{0026} % *** Enter the CVPR Paper ID here

\ifcvprfinal\fi
\begin{document}

	\title{Enhanced Deep Residual Networks for Single Image Super-Resolution}
	
	\author{
		Bee Lim \and Sanghyun Son \and Heewon Kim \and Seungjun Nah \and Kyoung Mu Lee \and\\
		Department of ECE, ASRI, Seoul National University, 08826, Seoul, Korea\\
		\tt\small forestrainee@gmail.com, thstkdgus35@snu.ac.kr, ghimhw@gmail.com\\
		\tt\small seungjun.nah@gmail.com, kyoungmu@snu.ac.kr
	}
	
	\maketitle

	\begin{abstract}
		
		Recent research on super-resolution has progressed with the development of deep convolutional neural networks (DCNN). 
		In particular, residual learning techniques exhibit improved performance. 
		In this paper, we develop an enhanced deep super-resolution network (\textbf{EDSR}) with performance exceeding those of current state-of-the-art SR methods. 
		The significant performance improvement of our model is due to optimization by removing unnecessary modules in conventional residual networks. 
		The performance is further improved by expanding the model size while we stabilize the training procedure.
		We also propose a new multi-scale deep super-resolution system (\textbf{MDSR}) and training method, which can reconstruct high-resolution images of different upscaling factors in a single model. 
		The proposed methods show superior performance over the state-of-the-art methods on benchmark datasets and prove its excellence by winning the NTIRE2017 Super-Resolution Challenge \cite{Timofte_2017_CVPR_Workshops}. 

	\end{abstract}

	\section{Introduction}
	
	Image super-resolution (SR) problem, particularly single image super-resolution (SISR), has gained increasing research attention for decades.
	SISR aims to reconstruct a high-resolution image $I^{SR}$ from a single low-resolution image $I^{LR}$. 
	Generally, the relationship between $I^{LR}$ and the original high-resolution image $I^{HR}$ can vary depending on the situation. 
	Many studies assume that $I^{LR}$ is a bicubic downsampled version of $I^{HR}$, but other degrading factors such as blur, decimation, or noise can also be considered for practical applications.
	
	\begin{figure}[t]
		\captionsetup[subfloat]{labelformat=empty}
		\begin{center}
			\newcommand{\rowArg}{2.65cm}
			\newcommand{\fullSize}{6.05cm}
			\newcommand{\patchSize}{2.6cm}
			% 			\begin{adjustbox}{width=\linewidth, center=\linewidth}
			\setlength\tabcolsep{0.05cm}
			\begin{tabular}[b]{c c c}
				\multicolumn{2}{c}{\multirow{2}{*}[\rowArg]{
						\subfloat[0853 from DIV2K~\cite{Timofte_2017_CVPR_Workshops}]
						% 			{\includegraphics[width = \fullSize, height = \fullSize]
						{\includegraphics[height=\fullSize, trim={0.7cm 0 0.7cm 0}, clip]
							{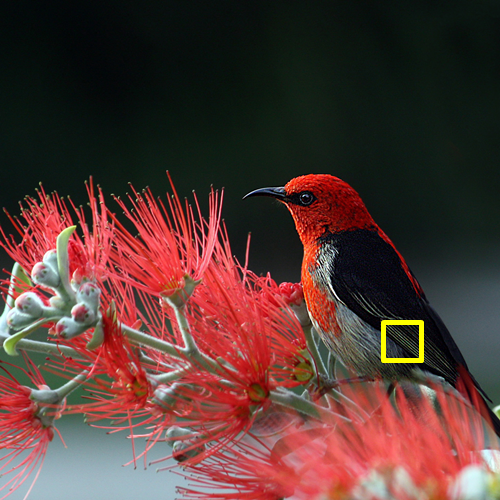}}}} &
				\subfloat[HR  \protect\linebreak(PSNR / SSIM)]{
					\includegraphics[width = \patchSize, height = \patchSize]
					{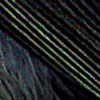}} \\ [-0.3cm]& &
				\subfloat[Bicubic  \protect\linebreak(30.80 dB / 0.9537)]{
					\includegraphics[width = \patchSize, height = \patchSize]
					{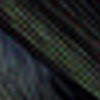}} \\ [-0.3cm]
				
				\subfloat[VDSR \cite{kim2016accurate}  \protect\linebreak(32.82 dB / 0.9623)]{
					\includegraphics[width = \patchSize, height = \patchSize]
					{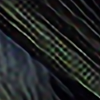}} &
				\subfloat[SRResNet \cite{ledig2016photo}  \protect\linebreak(34.00 dB / 0.9679)]{
					\includegraphics[width = \patchSize, height = \patchSize]
					{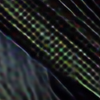}}	&
				\subfloat[\textbf{EDSR+} (Ours)  \protect\linebreak(\textcolor{red}{34.78 dB} / \textcolor{red}{0.9708})]{
					\includegraphics[width = \patchSize, height = \patchSize]
					{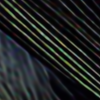}}		
			\end{tabular}		
			% 			\end{adjustbox}
		\end{center}
		\setlength{\abovecaptionskip}{0cm}
		\captionsetup{justification=raggedright,singlelinecheck=false}
		\caption{$\times 4$ Super-resolution result of our single-scale SR method (EDSR) compared with existing algorithms.}
		
	\end{figure}
	
	Recently, deep neural networks ~\cite{kim2016accurate,kim2016deeply, ledig2016photo} provide significantly improved performance in terms of peak signal-to-noise ratio (PSNR) in the SR problem. 
	However, such networks exhibit limitations in terms of architecture optimality. 
	First, the reconstruction performance of the neural network models is sensitive to minor architectural changes. 
	Also, the same model achieves different levels of performance by different initialization and training techniques. 
	Thus, carefully designed model architecture and sophisticated optimization methods are essential in training the neural networks.
	
	Second, most existing SR algorithms treat super-resolution of different scale factors as independent problems without considering and utilizing mutual relationships among different scales in SR.
	As such, those algorithms require many scale-specific networks that need to to be trained independently to deal with various scales. 
	Exceptionally, VDSR ~\cite{kim2016accurate} can handle super-resolution of several scales jointly in the single network. Training the VDSR model with multiple scales boosts the performance substantially and outperforms scale-specific training, implying the redundancy among scale-specific models.
	Nonetheless, VDSR style architecture requires bicubic interpolated image as the input, that leads to heavier computation time and memory compared to the architectures with scale-specific upsampling method~\cite{dong2016accelerating,shi2016real,ledig2016photo}.
	
	While SRResNet ~\cite{ledig2016photo} successfully solved those time and memory issue with good performance, it simply employs the ResNet architecture from He et al. [9] without much modification. However, original ResNet was proposed to solve higher-level computer vision problems such as image classification and detection. Therefore, applying ResNet architecture directly to low-level vision problems like super-resolution can be suboptimal.
		
	To solve these problems, based on the SRResNet architecture, we first optimize it by analyzing and removing unnecessary modules to simplify the network architecture. Training a network becomes nontrivial when the model is complex. Thus, we train the network with appropriate loss function and careful model modification upon training. We experimentally show that the modified scheme produces better results. 
	
	Second, we investigate the model training method that transfers knowledge from a model trained at other scales. To utilize scale-independent information during training, we train high-scale models from pre-trained low-scale models. Furthermore, we propose a new multi-scale architecture that shares most of the parameters across different scales. The proposed multi-scale model uses significantly fewer parameters compared with multiple single-scale models but shows comparable performance.
	
	We evaluate our models on the standard benchmark datasets and on a newly provided DIV2K dataset. The proposed single- and multi-scale super-resolution networks show the state-of-the-art performances on all datasets in terms of PSNR and SSIM. Our methods ranked first and second, respectively, in the NTIRE 2017 Super-Resolution Challenge~\cite{Timofte_2017_CVPR_Workshops}.

	\section{Related Works}
	
	To solve the super-resolution problem, early approaches use interpolation techniques based on sampling theory~\cite{allebach1996edge, li2001new, zhang2006edge}. However, those methods exhibit limitations in predicting detailed, realistic textures.
	Previous studies~\cite{tai2010super, sun2008image} adopted natural image statistics to the problem to reconstruct better high-resolution images.
	
	Advanced works aim to learn mapping functions between $I^{LR}$ and  $I^{HR}$ image pairs.
	Those learning methods rely on techniques ranging from neighbor embedding~\cite{chang2004super, bevilacqua2012low, gao2012image, roweis2000nonlinear}  to sparse coding~\cite{yang2012coupled, yang2010image, timofte2014a+, zeyde2010single}.
	Yang et al.~\cite{yang2013fast} introduced another approach that clusters the patch spaces and learns the corresponding functions.
	Some approaches utilize image self-similarities to avoid using external databases~\cite{glasner2009super, freedman2011image, wang2015learning}, and
	increase the size of the limited internal dictionary by geometric transformation of patches  ~\cite{huang2015single}.
	
	Recently, the powerful capability of deep neural networks has led to dramatic improvements in SR.
	Since Dong et al.~\cite{dong2014learning, dong2016accelerating} first proposed a deep learning-based SR method, various CNN architectures have been studied for SR.
	Kim et al.~\cite{kim2016accurate, kim2016deeply} first introduced the residual network for training much deeper network architectures and achieved superior performance. In particular, they showed that skip-connection and recursive convolution alleviate the burden of carrying identity information in the super-resolution network.
	Similarly to ~\cite{ronneberger2015u}, Mao et al.~\cite{mao2016image} tackled the general image restoration problem with encoder-decoder networks and symmetric skip connections. In ~\cite{mao2016image}, they argue that those nested skip connections provide fast and improved convergence.
	
	In many deep learning based super-resolution algorithms, an input image is upsampled via bicubic interpolation before they fed into the network ~\cite{dong2014learning,kim2016accurate,kim2016deeply}. Rather than using an interpolated image as an input, training upsampling modules at the very end of the network is also possible as shown in ~\cite{dong2016accelerating, shi2016real, ledig2016photo}. By doing so, one can reduce much of computations without losing model capacity because the size of features decreases. However, those kinds of approaches have one disadvantage: They cannot deal with the multi-scale problem in a single framework as in VDSR~\cite{kim2016accurate}.
	In this work, we resolve the dilemma of multi-scale training and computational efficiency. 
	We not only exploit the inter-relation of learned feature for each scale but also propose a new multi-scale model that efficiently reconstructs high-resolution images for various scales. 
	Furthermore, we develop an appropriate training method that uses multiple scales for both single- and multi-scale models. 
	
	Several studies also have focused on the loss functions to better train network models. 
	Mean squared error (MSE) or L2 loss is the most widely used loss function for general image restoration and is also major performance measure (PSNR) for those problems.
	However, Zhao et al. ~\cite{zhao2015loss} reported that training with L2 loss does not guarantee better performance compared to other loss functions in terms of PSNR and SSIM.
	In their experiments, a network trained with L1 achieved improved performance compared with the network trained with L2.
	
	\section{Proposed Methods}
	
	In this section, we describe proposed model architectures. 
	We first analyze recently published super-resolution network and suggest an enhanced version of the residual network architecture with the simpler structure. 
	We show that our network outperforms the original ones while exhibiting improved computational efficiency. 
	In the following sections, we suggest a single-scale architecture (EDSR) that handles a specific super-resolution scale and a multi-scale architecture (MDSR) that reconstructs various scales of high-resolution images in a single model.
	
	\subsection{Residual blocks}
	\label{sec_3_resnet}
	
	Recently, residual networks~\cite{kim2016accurate,he2016deep,ledig2016photo} exhibit excellent performance in computer vision problems from the low-level to high-level tasks. 
	Although Ledig et al.~\cite{ledig2016photo} successfully applied the ResNet architecture to the super-resolution problem with SRResNet, we further improve the performance by employing better ResNet structure.
	
	\begin{figure}[h]
		\begin{center}
			\setlength\tabcolsep{0.3cm}
			\begin{tabular}[b]{c c c}		
				\subfloat[Original]{\includegraphics[height=5cm]{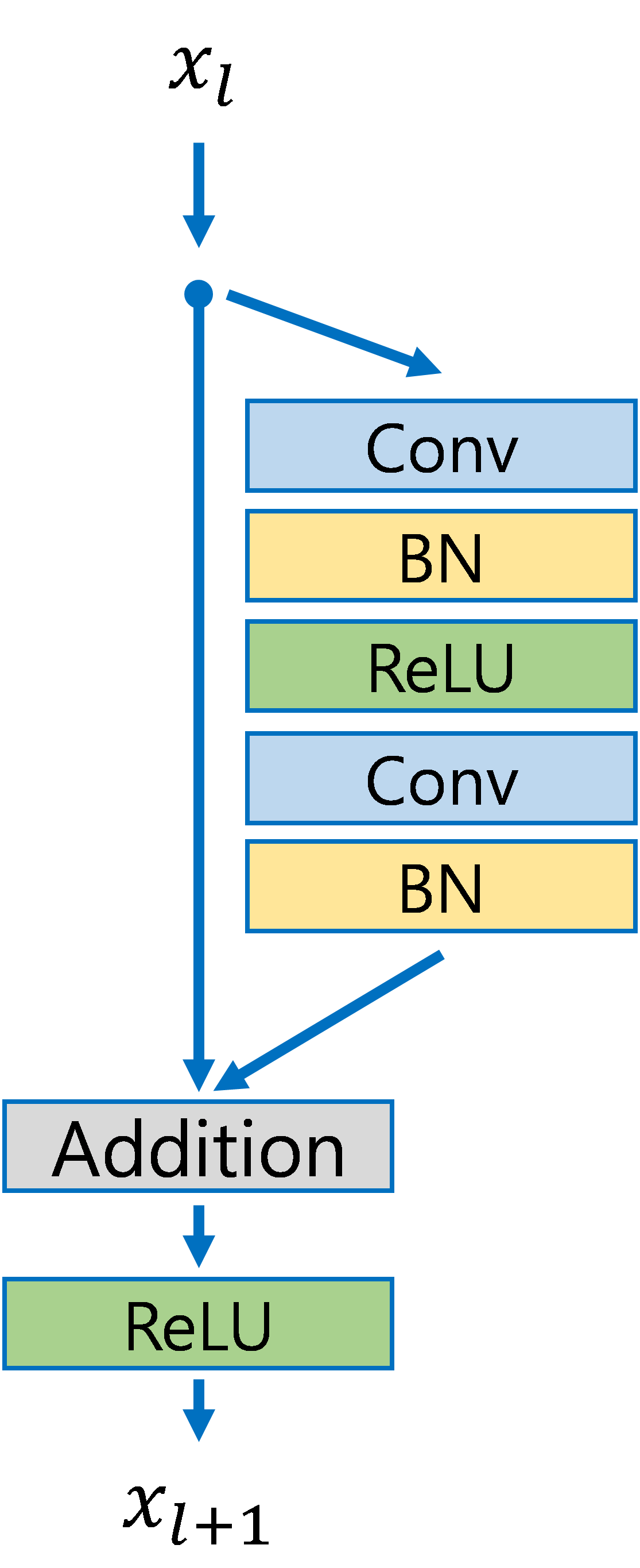}} &
				\subfloat[SRResNet]{\includegraphics[height=5cm]{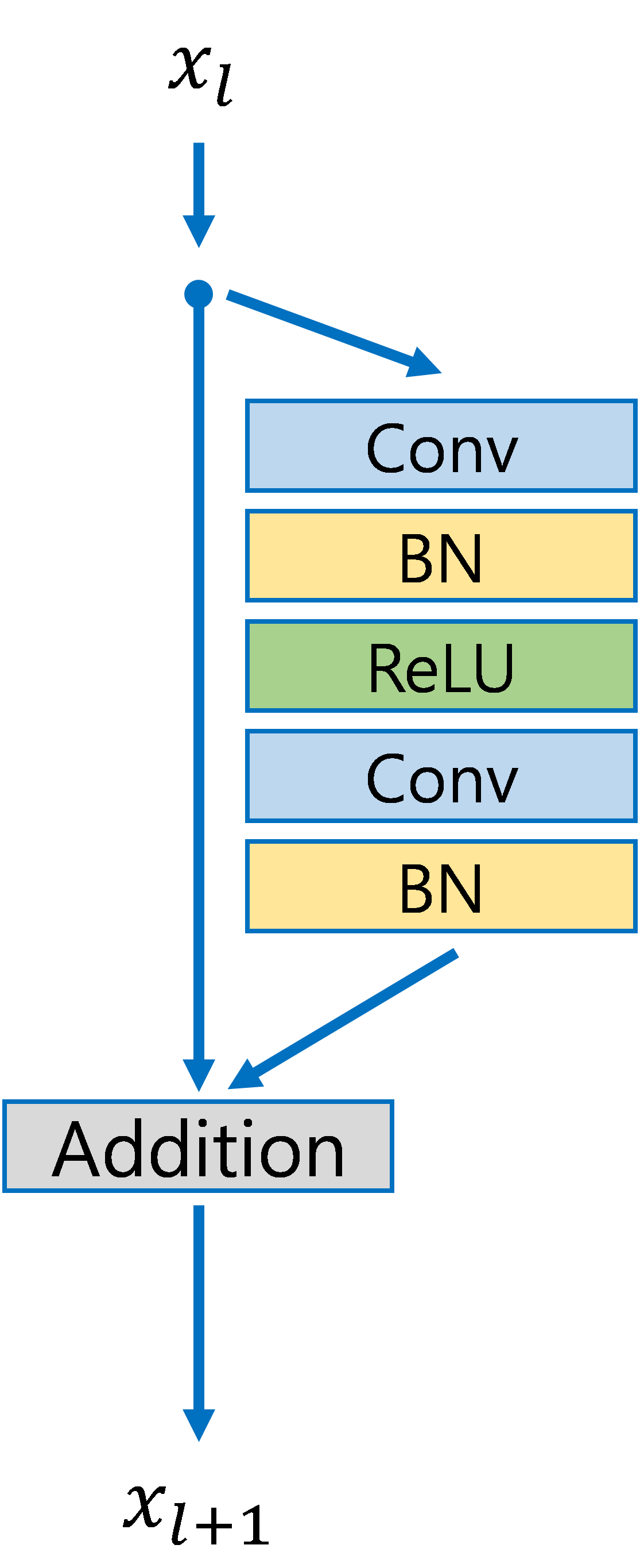}} &
				\subfloat[Proposed]{\includegraphics[height=5cm]{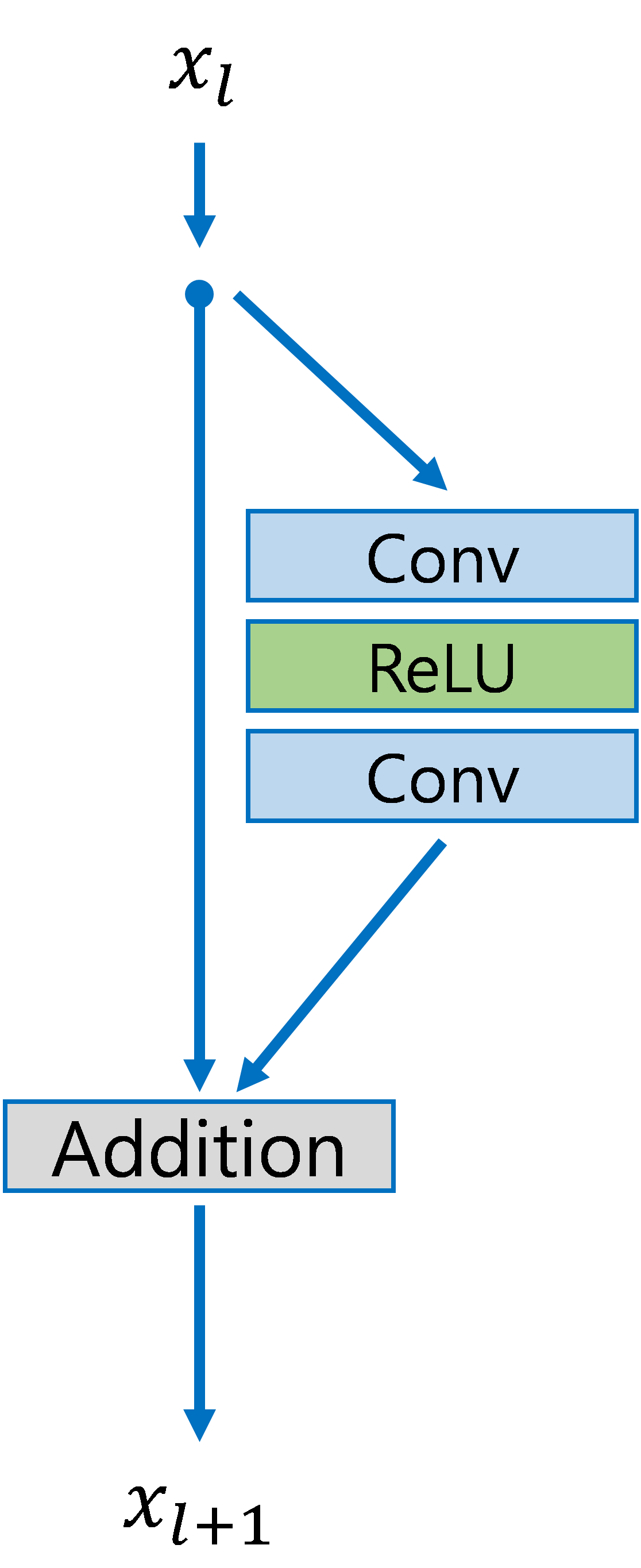}}
			\end{tabular}
		\end{center}
		\vspace*{-0.5cm}
		\captionsetup{justification=raggedright,singlelinecheck=false}
		\caption{Comparison of residual blocks in original ResNet, SRResNet, and ours.}
		\label{fig_resblock}
	\end{figure}
	
	In Fig.~\ref{fig_resblock}, we compare the building blocks of each network model from original ResNet~\cite{he2016deep}, SRResNet~\cite{ledig2016photo}, and our proposed networks. 
	We remove the batch normalization layers from our network as Nah et al.\cite{nah2016deep} presented in their image deblurring work. 
	Since batch normalization layers normalize the features, they get rid of range flexibility from networks by normalizing the features, it is better to remove them.
	We experimentally show that this simple modification increases the performance substantially as detailed in Sec. \ref{sec_experiments}.
	
	Furthermore, GPU memory usage is also sufficiently reduced since the batch normalization layers consume the same amount of memory as the preceding convolutional layers.
	Our baseline model without batch normalization layer saves approximately $40\%$ of memory usage during training, compared to SRResNet.
	Consequently, we can build up a larger model that has better performance than conventional ResNet structure under limited computational resources.
	
	% Single-scale architecture
	\begin{figure}[t]
		\begin{center}
			\includegraphics[width=\linewidth]{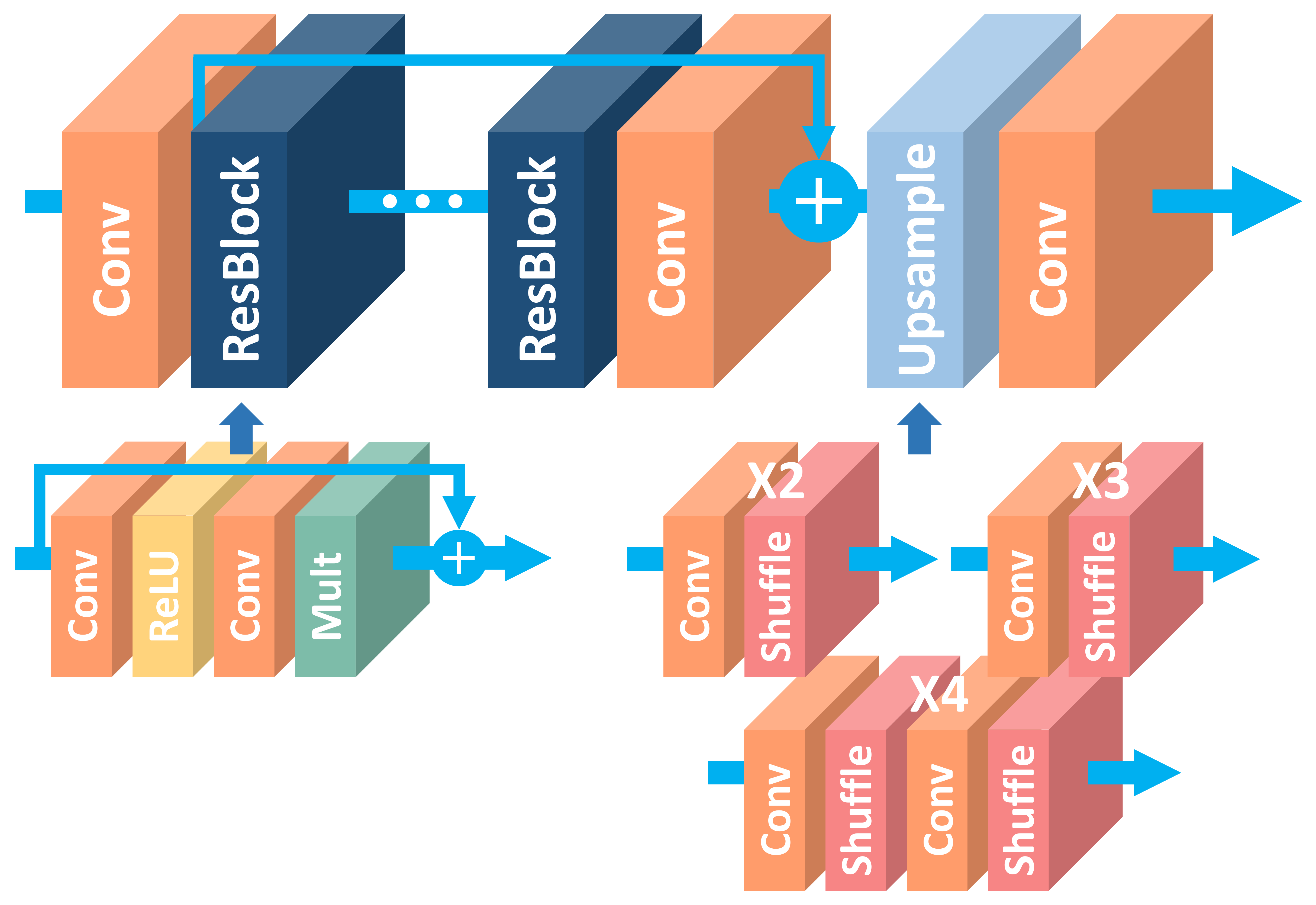}
		\end{center}
		\vspace*{-0.5cm}
		\captionsetup{justification=raggedright,singlelinecheck=false}
		\caption{The architecture of the proposed single-scale SR network (EDSR).}
		\label{fig_model_single}
	\end{figure}
	
	\subsection{Single-scale model} 
	\label{sec_3_single}
	
	The simplest way to enhance the performance of the network model is to increase the number of parameters.
	In the convolutional neural network, model performance can be enhanced by stacking many layers or by increasing the number of filters. 
	General CNN architecture with depth (the number of layers) $B$ and width (the number of feature channels) $F$ occupies roughly $O(BF)$ memory with $O(BF^2)$ parameters. 
	Therefore, increasing $F$ instead of $B$ can maximize the model capacity when considering limited computational resources.
	
	However, we found that increasing the number of feature maps above a certain level would make the training procedure numerically unstable. A similar phenomenon was reported by Szegedy et al.~\cite{szegedy2016inception}.
	We resolve this issue by adopting the residual scaling~\cite{szegedy2016inception} with factor 0.1.
	In each residual block, constant scaling layers are placed after the last convolution layers.
	These modules stabilize the training procedure greatly when using a large number of filters.
	In the test phase, this layer can be integrated into the previous convolution layer for the computational efficiency.
	
	We construct our \textbf{baseline (single-scale)} model with our proposed residual blocks in Fig.~\ref{fig_resblock}.
	The structure is similar to SRResNet~\cite{ledig2016photo}, but our model does not have ReLU activation layers outside the residual blocks.
	Also, our baseline model does not have residual scaling layers because we use only 64 feature maps for each convolution layer.
	In our final single-scale model (\textbf{EDSR}), we expand the baseline model by setting $B=32$, $F=256$ with a scaling factor 0.1.
	The model architecture is displayed in Fig.~\ref{fig_model_single}.
	
\begin{figure}[h]
		\begin{center}
			\includegraphics[width=\linewidth]{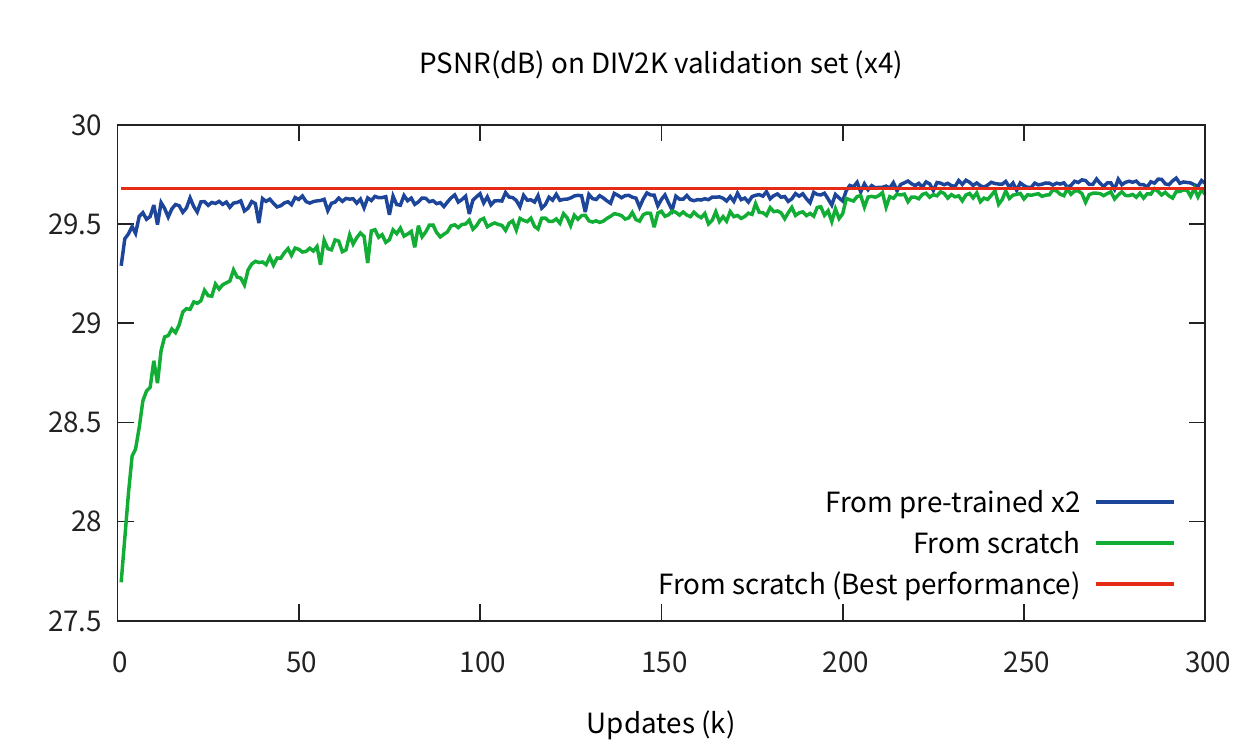}
		\end{center}
		\captionsetup{justification=raggedright,singlelinecheck=false}
		\caption{Effect of using pre-trained $\times 2$ network for $\times 4$ model (EDSR). The red line indicates the best performance of green line. 10 images are used for validation during training.}
		\label{fig_pretrained}
	\end{figure}

	When training our model for upsampling factor $\times 3$ and $\times 4$, we initialize the model parameters with pre-trained $\times2$ network.
	This pre-training strategy accelerates the training and improves the final performance as clearly demonstrated in Fig.~\ref{fig_pretrained}.
	For upscaling $\times 4$, if we use a pre-trained scale $\times 2$ model (blue line), the training converges much faster than the one started from random initialization (green line).
		
	% Effect of pre-traininig	
	
	\subsection{Multi-scale model}
	\label{sec_3_multi}    
	
\begin{figure}[t]
		\begin{center}
			\includegraphics[width=\linewidth]{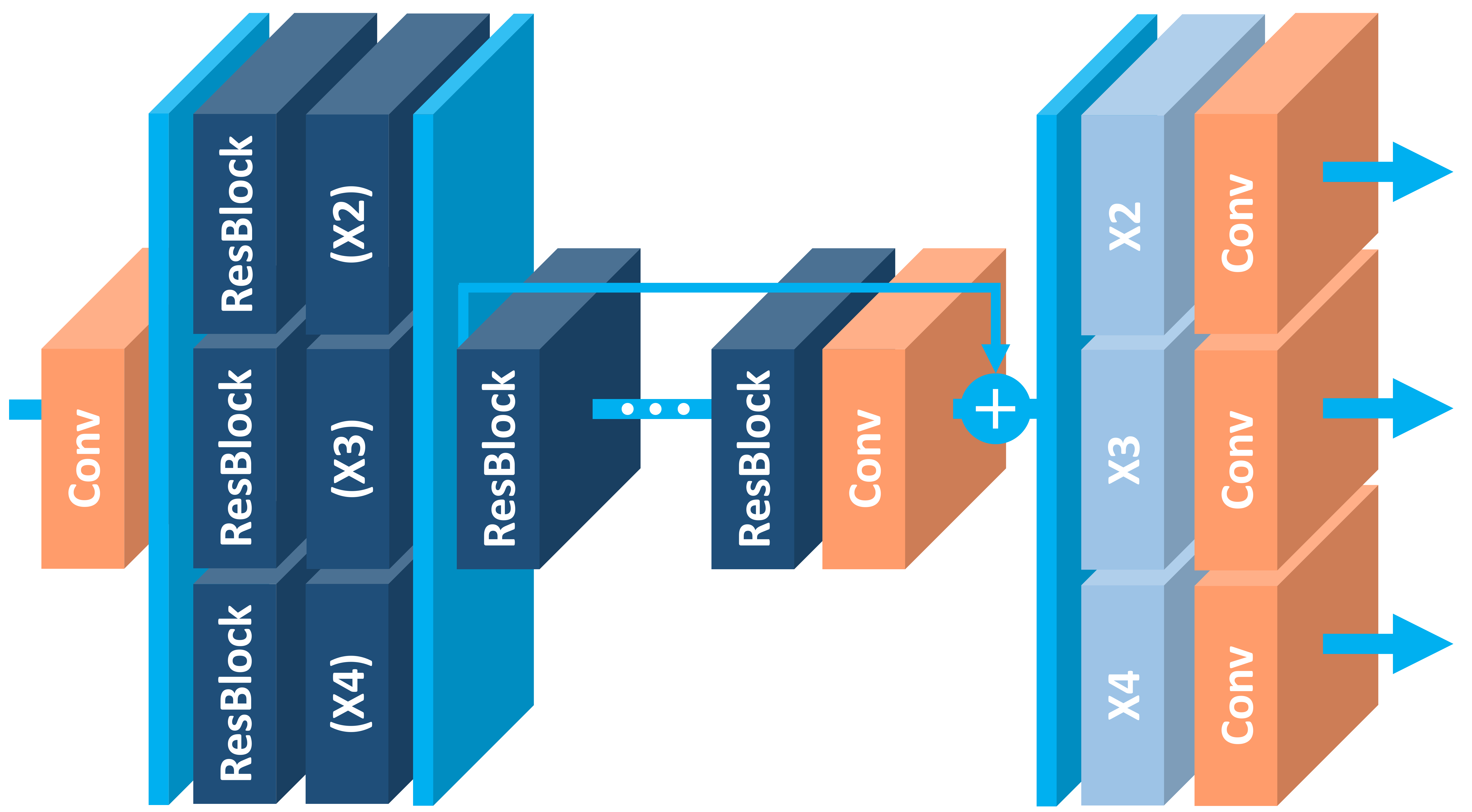}
		\end{center}
		% 		\vspace*{-0.5cm}
		\captionsetup{justification=raggedright,singlelinecheck=false}
		\caption{The architecture of the proposed multi-scale SR network (MDSR).}
		\label{fig_model_multi}
	\end{figure}

	From the observation in Fig.~\ref{fig_pretrained}, we conclude that super-resolution at multiple scales is inter-related tasks.
	We further explore this idea by building a multi-scale architecture that takes the advantage of inter-scale correlation as VDSR \cite{kim2016accurate} does.
	We design our \textbf{baseline (multi-scale)} models to have a single main branch with $B=16$ residual blocks so that most of the parameters are shared across different scales as shown in Fig.~\ref{fig_model_multi}.
	
	In our multi-scale architecture, we introduce scale-specific processing modules to handle the super-resolution at multiple scales.
	First, pre-processing modules are located at the head of networks to reduce the variance from input images of different scales.
	Each of pre-processing module consists of two residual blocks with $5 \times 5$ kernels.
	By adopting larger kernels for pre-processing modules, we can keep the scale-specific part shallow while the larger receptive field is covered in early stages of networks.
	At the end of the multi-scale model, scale-specific upsampling modules are located in parallel to handle multi-scale reconstruction.
	The architecture of the upsampling modules is similar to those of single-scale models described in the previous section.
	
	We construct our final multi-scale model (\textbf{MDSR}) with $B=80$ and $F=64$.
	While our single-scale baseline models for 3 different scales have about 1.5M parameters each, totaling 4.5M, our baseline multi-scale model has only 3.2 million parameters. Nevertheless, the multi-scale model exhibits comparable performance as the single-scale models. Furthermore, our multi-scale model is scalable in terms of depth. Although our final MDSR has approximately 5 times more depth compared to the baseline multi-scale model, only 2.5 times more parameters are required, as the residual blocks are lighter than scale-specific parts. Note that MDSR also shows the comparable performance to the scale-specific EDSRs. The detailed performance comparison of our proposed models is presented in Table~\ref{table_div2k_300k} and~\ref{table_bench}.
	
	% Multi-scale architecture
	
	% Model spec Table
	\begin{table}[t]
		{\footnotesize
			\renewcommand{\arraystretch}{1.3}
			\setlength{\tabcolsep}{2pt}
			\begin{center}
				\begin{tabular}{ c | c c c c }
					\\ [-1em]
					Options 
					& \parbox[c]{2cm} {\centering SRResNet~\cite{ledig2016photo} \\(reproduced)}
					& \parbox[c]{2cm} {\centering Baseline \\ (Single / Multi)}
					& EDSR
					& MDSR\\ \\ [-1em]\hline
					$\#$ Residual blocks & 16 & 16 & 32 & 80\\ % \hline
					$\#$ Filters & 64 & 64 & 256 & 64\\ % \hline
					$\#$ Parameters & 1.5M & 1.5M / 3.2M & 43M & 8.0M\\ % \hline
					Residual scaling & - & - & 0.1 & -\\ % \hline
					Use BN & Yes & No & No & No\\
					Loss function & L2 & L1 & L1 & L1\\ % \hline
					%Pre-training & No & Yes & Yes & Yes\\ % \hline
				\end{tabular}
			\end{center}
		}
		\caption{Model specifications.}
		\label{table_models}
	\end{table}
	
	\section{Experiments}
	\label{sec_experiments}
	
	\subsection{Datasets}
	
	DIV2K dataset~\cite{Timofte_2017_CVPR_Workshops} is a newly proposed high-quality (2K resolution) image dataset for image restoration tasks. 
	The DIV2K dataset consists of 800 training images, 100 validation images, and 100 test images.
	As the test dataset ground truth is not released, we report and compare the performances on the validation dataset.
	%We train our models in the training images of the DIV2K dataset and report the validation performance for comparison.
	We also compare the performance on four standard benchmark datasets: Set5~\cite{bevilacqua2012low}, Set14~\cite{zeyde2010single}, B100~\cite{martin2001database}, and Urban100~\cite{huang2015single}.
	
	\subsection{Training Details} 
	\label{sec_training_details}
	
	% Comparison with SRResNet
	\begin{table*}[h]
		{\scriptsize
			\renewcommand{\arraystretch}{1.4}
			\newcommand{\colWidth}{1.6cm}
			\setlength\tabcolsep{0.1cm}
			\begin{center}
				\begin{tabular}{ |c|c|c|c|c|c|c|c|c| }
					\hline &&&&&&&& \\ [-1em]
					% Scale & \parbox[c]{1.5cm}{\centering SRResNet\\(L2)} 
					%     & \parbox[c]{1.5cm}{\centering SRResNet\\(L1)}
					% 	& \parbox[c]{1.6cm}{\centering Our baseline\\(Single-scale)}
					% 	& \parbox[c]{1.5cm}{\centering Our baseline\\(Multi-scale)} \\ \\ [-1em]\hline
					Scale & \parbox[c]{\colWidth}{\centering SRResNet\\ (L2 loss)}
					& \parbox[c]{\colWidth}{\centering SRResNet\\ (L1 loss)}
					& \parbox[c]{\colWidth}{\centering Our baseline\\ (Single-scale)}
					& \parbox[c]{\colWidth}{\centering Our baseline\\ (Multi-scale)} 
					& \parbox[c]{1.5cm}{\centering EDSR\\ (Ours)}
					& \parbox[c]{1.5cm}{\centering MDSR\\ (Ours)}
					& \parbox[c]{1.5cm}{\centering\textbf{EDSR+\\ (Ours)}} 
					& \parbox[c]{1.5cm}{\centering\textbf{MDSR+\\ (Ours)}} \\ 
					&&&&&&&& \\ [-1em]\hline
					$\times 2$ &
					34.40 / 0.9662 &
					34.44 / 0.9665 &
					34.55 / 0.9671 &
					34.60 / 0.9673 &
					35.03 / 0.9695 &	
					34.96 / 0.9692 &
					\textcolor{red}{35.12} / \textcolor{red}{0.9699} &
					\textcolor{blue}{35.05} / \textcolor{blue}{0.9696} \\
					$\times 3$ &
					30.82 / 0.9288 &
					30.85 / 0.9292 &
					30.90 / 0.9298 &
					30.91 / 0.9298 &
					31.26 / 0.9340 &	
					31.25 / 0.9338 &
					\textcolor{red}{31.39} / \textcolor{red}{0.9351} &
					\textcolor{blue}{31.36} / \textcolor{blue}{0.9346} \\
					$\times 4$ &
					28.92 / 0.8960 &
					28.92 / 0.8961 &
					28.94 / 0.8963 &
					28.95 / 0.8962 &
					29.25 / 0.9017 &
					29.26 / 0.9016 &
					\textcolor{red}{29.38} / \textcolor{red}{0.9032} &
					\textcolor{blue}{29.36} / \textcolor{blue}{0.9029} \\
					\hline
				\end{tabular}
			\end{center}
		}
		\captionsetup{justification=raggedright,singlelinecheck=false}
		\caption{Performance comparison between architectures on the DIV2K validation set (PSNR(dB) / SSIM). Red indicates the best performance and blue indicates the second best. EDSR+ and MDSR+ denote self-ensemble versions of EDSR and MDSR.}
		\label{table_div2k_300k}
	\end{table*}
	
	For training, we use the RGB input patches of size $48\times 48$ from LR image with the corresponding HR patches.
	We augment the training data with random horizontal flips and 90° rotations.
	We pre-process all the images by subtracting the mean RGB value of the DIV2K dataset.
	We train our model with ADAM optimizer~\cite{kingma2014adam} by setting $\beta_1 = 0.9$, $\beta_2 = 0.999$, and $\epsilon = 10^{-8}$.
	We set minibatch size as 16.
	The learning rate is initialized as $10^{-4}$ and halved at every $2\times10^{5}$ minibatch updates.
	
	For the single-scale models (EDSR), we train the networks as described in Sec.~\ref{sec_3_single}.
	The $\times 2$ model is trained from scratch.
	After the model converges, we use it as a pre-trained network for other scales.
	
	At each update of training a multi-scale model (MDSR), we construct the minibatch with a randomly selected scale among ${\times 2, \times 3}$ and ${\times 4}$.
	Only the modules that correspond to the selected scale are enabled and updated.
	Hence, scale-specific residual blocks and upsampling modules that correspond to different scales other than the selected one are not enabled nor updated.
	
	We train our networks using L1 loss instead of L2. Minimizing L2 is generally preferred since it maximizes the PSNR. However, based on a series of experiments we empirically found that L1 loss provides better convergence than L2. The evaluation of this comparison is provided in Sec. \ref{sec_compare}
	
	We implemented the proposed networks with the Torch7 framework and trained them using NVIDIA Titan X GPUs. It takes 8 days and 4 days to train EDSR and MDSR, respectively.
	The source code is publicly available online.\footnote{https://github.com/LimBee/NTIRE2017}
	
	\subsection{Geometric Self-ensemble}
	\label{sec_ensemble}
	
	In order to maximize the potential performance of our model, we adopt the self-ensemble strategy similarly to  \cite{timofte2016seven}. During the test time, we flip and rotate the input image $I^{LR}$ to generate seven augmented inputs $I_{n,i}^{LR} = T_i \left( I_n^{LR} \right)$ for each sample, where $T_i$ represents the 8 geometric transformations including indentity.
	With those augmented low-resolution images, we generate corresponding super-resolved images $\left \{ I_{n,1}^{SR},\cdots,I_{n,8}^{SR}\right \}$ using the networks. We then apply inverse transform to those output images to get the original geometry $\tilde{I}_{n,i}^{SR} = T_i^{-1} \left( I_{n,i}^{SR} \right)$. Finally, we average the transformed outputs all together to make the self-ensemble result as follows. 
	$I_n^{SR}=\frac{1}{8} \sum \limits_{i=1}^{8} \tilde I_{n,i}^{SR}$.
	
	This self-ensemble method has an advantage over other ensembles as it does not require additional training of separate models. It is beneficial especially when the model size or training time matters. Although self-ensemble strategy keeps the total number of parameters same, we notice that it gives approximately same performance gain compared to conventional model ensemble method that requires individually trained models. We denote the methods using self-ensemble by adding '+' postfix to the method name; i.e. EDSR+/MDSR+. Note that geometric self-ensemble is valid only for symmetric downsampling methods such as bicubic downsampling. 
	
	% Qualitative results
	\begin{figure*}[t]
		\captionsetup[subfloat]{labelformat=empty}
		\begin{center}
			\newcommand{\rowArg}{2.6cm}
			\newcommand{\fullSize}{5.85cm}
			\newcommand{\patchSize}{2.45cm}
			\scriptsize
			\setlength\tabcolsep{0.1cm}
			\begin{tabular}[b]{c c c c c}
				\multirow{2}{*}[\rowArg]{
					\subfloat[img034 from Urban100~\cite{huang2015single}]
					{\includegraphics[width = \fullSize, height = \fullSize]
						{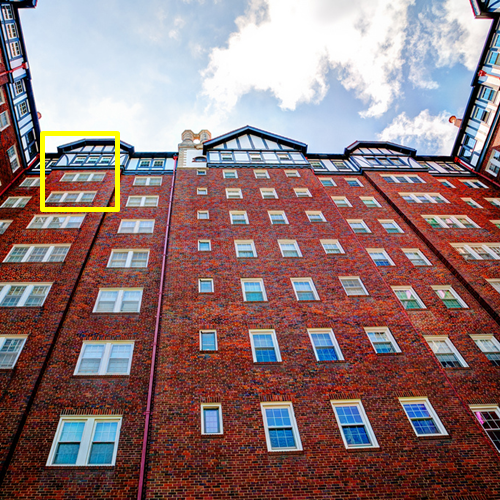}}} &
				\subfloat[HR \protect\linebreak(PSNR / SSIM)]
				{\includegraphics[width = \patchSize, height = \patchSize]
					{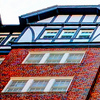}} &
				\subfloat[Bicubic \protect\linebreak(21.41 dB / 0.4810)]
				{\includegraphics[width = \patchSize, height = \patchSize]
					{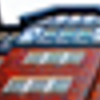}} &
				\subfloat[A+ \cite{timofte2014a+} \protect\linebreak(22.21 dB / 0.5408)]
				{\includegraphics[width = \patchSize, height = \patchSize]
					{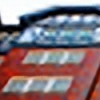}} &
				\subfloat[SRCNN \cite{dong2014learning} \protect\linebreak(22.33 dB / 0.5461)]
				{\includegraphics[width = \patchSize, height = \patchSize]
					{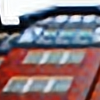}} \\ [-0.2cm] &
				\subfloat[VDSR \cite{kim2016accurate} \protect\linebreak(22.62 dB / 0.5657)]
				{\includegraphics[width = \patchSize, height = \patchSize]
					{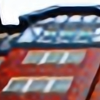}} &
				\subfloat[SRResNet \cite{ledig2016photo} \protect\linebreak(23.14 dB / 0.5891)]
				{\includegraphics[width = \patchSize, height = \patchSize]
					{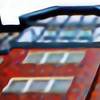}} &
				\subfloat[\textbf{EDSR+ (Ours)} \protect\linebreak(\textcolor{red}{23.48 dB} / \textcolor{red}{0.6048})]
				{\includegraphics[width = \patchSize, height = \patchSize]
					{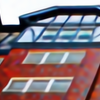}} &
				\subfloat[\textbf{MDSR+ (Ours)} \protect\linebreak(\textcolor{blue}{23.46 dB} / \textcolor{blue}{0.6039})]
				{\includegraphics[width = \patchSize, height = \patchSize]
					{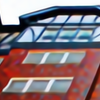}} \\
				
				\multirow{2}{*}[\rowArg]{
					\subfloat[img062 from Urban100~\cite{huang2015single}]
					{\includegraphics[width = \fullSize, height = \fullSize]
						{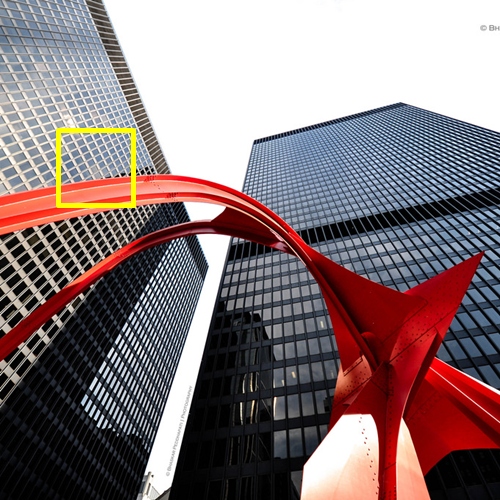}}} &
				\subfloat[HR \protect\linebreak(PSNR / SSIM)]
				{\includegraphics[width = \patchSize, height = \patchSize]
					{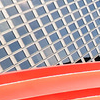}} &
				\subfloat[Bicubic \protect\linebreak(19.82 dB / 0.6471)]
				{\includegraphics[width = \patchSize, height = \patchSize]
					{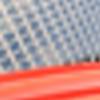}} &
				\subfloat[A+ \cite{timofte2014a+} \protect\linebreak(20.43 dB \ 0.7145)]
				{\includegraphics[width = \patchSize, height = \patchSize]
					{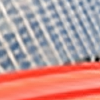}} &
				\subfloat[SRCNN \cite{dong2014learning} \protect\linebreak(20.61 dB / 0.7218)]
				{\includegraphics[width = \patchSize, height = \patchSize]
					{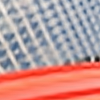}} \\ [-0.2cm] &
				\subfloat[VDSR \cite{kim2016accurate} \protect\linebreak(20.75 dB / 0.7504)]
				{\includegraphics[width = \patchSize, height = \patchSize]
					{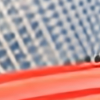}} &
				\subfloat[SRResNet \cite{ledig2016photo} \protect\linebreak(21.70 dB / 0.8054)]
				{\includegraphics[width = \patchSize, height = \patchSize]
					{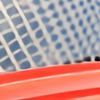}} &
				\subfloat[\textbf{EDSR+ (Ours)} \protect\linebreak(\textcolor{red}{22.70 dB} / \textcolor{red}{0.8537})]
				{\includegraphics[width = \patchSize, height = \patchSize]
					{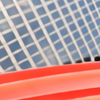}} &
				\subfloat[\textbf{MDSR+ (Ours)} \protect\linebreak(\textcolor{blue}{22.66 dB} / \textcolor{blue}{0.8508})]
				{\includegraphics[width = \patchSize, height = \patchSize]
					{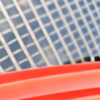}} \\
				
				\multirow{2}{*}[\rowArg]{
					\subfloat[0869 from DIV2K~\cite{Timofte_2017_CVPR_Workshops}]
					{\includegraphics[width = \fullSize, height = \fullSize]
						{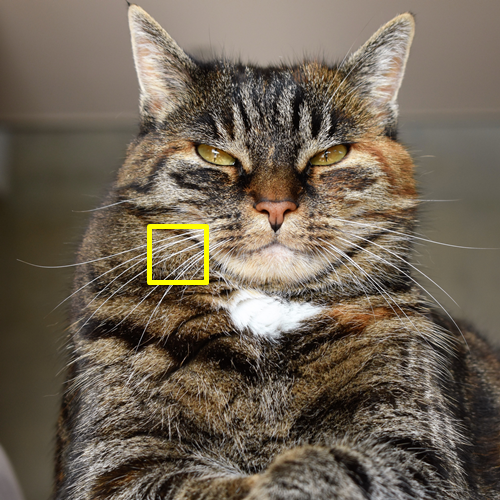}}} &
				\subfloat[HR \protect\linebreak(PSNR / SSIM)]
				{\includegraphics[width = \patchSize, height = \patchSize]
					{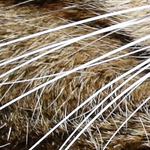}} &
				\subfloat[Bicubic \protect\linebreak(22.66 dB / 0.8025)]
				{\includegraphics[width = \patchSize, height = \patchSize]
					{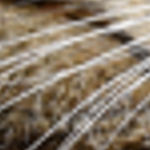}} &
				\subfloat[A+ \cite{timofte2014a+} \protect\linebreak(23.10 dB / 0.8251)]
				{\includegraphics[width = \patchSize, height = \patchSize]
					{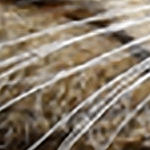}} &
				\subfloat[SRCNN \cite{dong2014learning} \protect\linebreak(23.14 dB / 0.8280)]
				{\includegraphics[width = \patchSize, height = \patchSize]
					{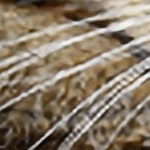}} \\ [-0.2cm] &
				\subfloat[VDSR \cite{kim2016accurate} \protect\linebreak(23.36 dB / 0.8365)]
				{\includegraphics[width = \patchSize, height = \patchSize]
					{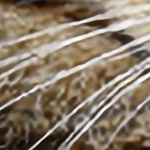}} &
				\subfloat[SRResNet \cite{ledig2016photo} \protect\linebreak(23.71 dB / 0.8485)] % \protect\footnotemark ]
				{\includegraphics[width = \patchSize, height = \patchSize]
					{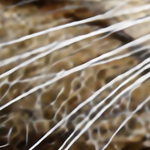}} &
				\subfloat[\textbf{EDSR+ (Ours)} \protect\linebreak(\textcolor{blue}{23.89 dB} / \textcolor{red}{0.8563})]
				{\includegraphics[width = \patchSize, height = \patchSize]
					{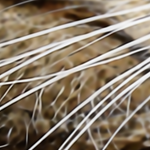}} &
				\subfloat[\textbf{MDSR+ (Ours)} \protect\linebreak(\textcolor{red}{23.90 dB} \textcolor{blue}{/ 0.8558})]
				{\includegraphics[width = \patchSize, height = \patchSize]
					{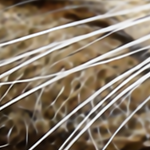}} \\
			\end{tabular}
		\end{center}
		\setlength{\abovecaptionskip}{0pt plus 2pt minus 2pt}
		\setlength{\belowcaptionskip}{0pt plus 2pt minus 2pt}
		%\captionsetup{justification=raggedright,singlelinecheck=false}
		\caption{Qualitative comparison of our models with other works on $\times 4$ super-resolution.}
		\label{fig_result_1}
		% 		\footnotetext{We used our reproduction of SRResNet for this image.}
	\end{figure*}
	
	\subsection{Evaluation on DIV2K Dataset}
	\label{sec_compare}
	
	We test our proposed networks on the DIV2K dataset. Starting from the SRResNet, we gradually change various settings to perform ablation tests. We train SRResNet \cite{ledig2016photo} on our own. 
	\footnote{We confirmed our reproduction is correct by getting comparable results in an individual experiment, using the same settings of the paper \cite{ledig2016photo}. In our experiments, however, it became slightly different to match the settings of our baseline model training. See our codes at https://github.com/LimBee/NTIRE2017.}
	\footnote{We used the original paper (https://arxiv.org/abs/1609.04802v3) as a reference.}
	First, we change the loss function from L2 to L1, and then the network architecture is reformed as described in the previous section and summarized in Table ~\ref{table_models}.
	
	We train all those models with $3\times10^{5}$ updates in this experiment. 
	Evaluation is conducted on the 10 images of DIV2K validation set, with PSNR and SSIM criteria. For the evaluation, we use full RGB channels and ignore the (6 + scale) pixels from the border.
	
	Table~\ref{table_div2k_300k} presents the quantitative results. SRResNet trained with L1 gives slightly better results than the original one trained with L2 for all scale factors. Modifications of the network give an even bigger margin of improvements. The last 2 columns of Table ~\ref{table_div2k_300k} show significant performance gains of our final bigger models, EDSR+ and MDSR+ with the geometric  self-ensemble technique.
	Note that our models require much less GPU memory since they do not have batch normalization layers.
	
	% Benchmark table
	\begin{table*}[th]
		{\scriptsize
			\renewcommand{\arraystretch}{1.4}
			\setlength\tabcolsep{5pt}
			\begin{center}
				\begin{adjustbox}{width=1.1\textwidth, center=\textwidth+0.5cm}
					\begin{tabular} {|*{11}{c|}}
						\hline 
						&&&&& &&&&& \\ [-1em]
						Dataset & Scale & Bicubic &
						A+ ~\cite{timofte2014a+}&
						SRCNN ~\cite{dong2014learning}&
						VDSR ~\cite{kim2016accurate}&
						SRResNet ~\cite{ledig2016photo}&
						\parbox[c]{1.5cm}{\centering EDSR \\(Ours)} &
						\parbox[c]{1.5cm}{\centering MDSR \\(Ours)} &
						\parbox[c]{1.5cm}{\centering\textbf{EDSR+ \\(Ours)}} &
						\parbox[c]{1.5cm}{\centering\textbf{MDSR+ \\(Ours)}} \\ 
						&&&&& &&&&&\\ [-1em]
						\hline
						& $\times 2$ &
						33.66 / 0.9299 & 36.54 / 0.9544 & 36.66 / 0.9542 & 37.53 / 0.9587 & - / - &
						38.11 / 0.9601 & 
						38.11 / 0.9602 &
						\textcolor{red}{38.20} / \textcolor{red}{0.9606} & 
						\textcolor{blue}{38.17} / \textcolor{blue}{0.9605} \\
						Set5 & $\times 3$ &
						30.39 / 0.8682 & 32.58 / 0.9088 & 32.75 / 0.9090 & 33.66 / 0.9213 & - / - &
						34.65 / 0.9282 & 
						34.66 / 0.9280 &
						\textcolor{blue}{34.76} / \textcolor{red}{0.9290} &
						\textcolor{red}{34.77} / \textcolor{blue}{0.9288} \\
						& $\times 4$ &
						28.42 / 0.8104 & 30.28 / 0.8603 & 30.48 / 0.8628 & 31.35 / 0.8838 & 32.05 / 
						0.8910 & 
						32.46 / 0.8968 & 
						32.50 / 0.8973 &
						\textcolor{red}{32.62} / \textcolor{red}{0.8984} & 
						\textcolor{blue}{32.60} / \textcolor{blue}{0.8982} \\
						\hline
						& $\times 2$ &
						30.24 / 0.8688 & 32.28 / 0.9056 & 32.42 / 0.9063 & 33.03 / 0.9124 & - / - &
						33.92 / 0.9195 & 
						33.85 / 0.9198 &
						\textcolor{red}{34.02} / \textcolor{red}{0.9204} & 
						\textcolor{blue}{33.92} / \textcolor{blue}{0.9203} \\
						Set14 & $\times 3$ &
						27.55 / 0.7742 & 29.13 / 0.8188 & 29.28 / 0.8209 & 29.77 / 0.8314 & - / - &
						30.52 / 0.8462 & 
						30.44 / 0.8452 &
						\textcolor{red}{30.66} / \textcolor{red}{0.8481} & 
						\textcolor{blue}{30.53} / \textcolor{blue}{0.8465} \\
						& $\times 4$ &
						26.00 / 0.7027 & 27.32 / 0.7491 & 27.49 / 0.7503 & 28.01 / 0.7674 & 
						28.53 / 0.7804 &
						28.80 / 0.7876 & 
						28.72 / 0.7857 &
						\textcolor{red}{28.94} / \textcolor{red}{0.7901} & 
						\textcolor{blue}{28.82} / \textcolor{blue}{0.7876} \\
						\hline
						& $\times 2$ &
						29.56 / 0.8431 & 31.21 / 0.8863 & 31.36 / 0.8879 & 31.90 / 0.8960 & - / - &
						32.32 / 0.9013 & 
						32.29 / 0.9007 &
						\textcolor{red}{32.37} / \textcolor{red}{0.9018} & 
						\textcolor{blue}{32.34} / \textcolor{blue}{0.9014} \\						
						B100 & $\times 3$ &
						27.21 / 0.7385 & 28.29 / 0.7835 & 28.41 / 0.7863 & 28.82 / 0.7976 & - / - &
						29.25 / 0.8093 & 
						29.25 / 0.8091 &
						\textcolor{red}{29.32} / \textcolor{red}{0.8104} & 
						\textcolor{blue}{29.30} / \textcolor{blue}{0.8101} \\
						& $\times 4$ &
						25.96 / 0.6675 & 26.82 / 0.7087 & 26.90 / 0.7101 & 27.29 / 0.7251 & 
						27.57 / 0.7354 &
						27.71 / 0.7420 & 
						27.72 / 0.7418 &
						\textcolor{red}{27.79} / \textcolor{red}{0.7437} & 
						\textcolor{blue}{27.78} / \textcolor{blue}{0.7425} \\
						\hline
						& $\times 2$ &
						26.88 / 0.8403 & 29.20 / 0.8938 & 29.50 / 0.8946 & 30.76 / 0.9140 & - / - &
						32.93 / 0.9351 & 
						32.84 / 0.9347 &
						\textcolor{red}{33.10} / \textcolor{red}{0.9363} & 
						\textcolor{blue}{33.03} / \textcolor{blue}{0.9362} \\
						Urban100 & $\times 3$ &
						24.46 / 0.7349 & 26.03 / 0.7973 & 26.24 / 0.7989 & 27.14 / 0.8279 & - / - &
						28.80 / 0.8653 & 
						28.79 / 0.8655 &
						\textcolor{red}{29.02} / \textcolor{red}{0.8685} & 
						\textcolor{blue}{28.99} / \textcolor{blue}{0.8683} \\
						& $\times 4$ &
						23.14 / 0.6577 & 24.32 / 0.7183 & 24.52 / 0.7221 & 25.18 / 0.7524 & 
						26.07 / 0.7839 & 
						26.64 / 0.8033 & 
						26.67 / 0.8041 &
						\textcolor{red}{26.86} / \textcolor{blue}{0.8080} & 
						\textcolor{red}{26.86} / \textcolor{red}{0.8082} \\
						\hline\hline
						\multirow{3}{*}{\parbox[c]{1cm}{\centering DIV2K\\validation}}
						& $\times 2$ &
						31.01 / 0.9393 & 32.89 / 0.9570 & 33.05 / 0.9581 & 33.66 / 0.9625 & - / - &
						35.03 / 0.9695 &	
						34.96 / 0.9692 &
						\textcolor{red}{35.12} / \textcolor{red}{0.9699} &
						\textcolor{blue}{35.05} / \textcolor{blue}{0.9696} \\
						& $\times 3$ &
						28.22 / 0.8906 & 29.50 / 0.9116 & 29.64 / 0.9138 & 30.09 / 0.9208 & - / - &
						31.26 / 0.9340 &	
						31.25 / 0.9338 &
						\textcolor{red}{31.39} / \textcolor{red}{0.9351} &
						\textcolor{blue}{31.36} / \textcolor{blue}{0.9346} \\
						& $\times 4$ &
						26.66 / 0.8521 & 27.70 / 0.8736 & 27.78 / 0.8753 & 28.17 / 0.8841 & - / - &
						29.25 / 0.9017 &
						29.26 / 0.9016 &
						\textcolor{red}{29.38} / \textcolor{red}{0.9032} &
						\textcolor{blue}{29.36} / \textcolor{blue}{0.9029} \\
						\hline								
					\end{tabular}
				\end{adjustbox}
			\end{center}
		}
		\captionsetup{justification=raggedright,singlelinecheck=false}
		\caption{Public benchmark test results and DIV2K validation results (PSNR(dB) / SSIM). 
			Red indicates the best performance and blue indicates the second best.
			Note that DIV2K validation results are acquired from published demo codes.}
		\label{table_bench}
	\end{table*}
	
	%DIV2K validation
	\iffalse
	\hline\hline
	\multirow{3}{*}{\parbox[c]{1cm}{\centering DIV2K\\validation}}
	& $\times 2$ &
	31.01 / 0.9393 & 32.89 / 0.9570 & 33.05 / 0.9581 & 33.66 / 0.9625 & - / - &
	35.03 / 0.9695 &	
	34.96 / 0.9692 &
	\textcolor{red}{35.12} / \textcolor{red}{0.9699} &
	\textcolor{blue}{35.05} / \textcolor{blue}{0.9696} \\
	& $\times 3$ &
	28.22 / 0.8906 & 29.50 / 0.9116 & 29.64 / 0.9138 & 30.09 / 0.9208 & - / - &
	31.26 / 0.9340 &	
	31.25 / 0.9338 &
	\textcolor{red}{31.39} / \textcolor{red}{0.9351} &
	\textcolor{blue}{31.36} / \textcolor{blue}{0.9346} \\
	& $\times 4$ &
	26.66 / 0.8521 & 27.70 / 0.8736 & 27.78 / 0.8753 & 28.17 / 0.8841 & - / - &
	29.25 / 0.9017 &
	29.26 / 0.9016 &
	\textcolor{red}{29.38} / \textcolor{red}{0.9032} &
	\textcolor{blue}{29.36} / \textcolor{blue}{0.9029} \\      					
	\fi
	
	\subsection{Benchmark Results}
	
	We provide the quantitative evaluation results of our final models (EDSR+, MDSR+) on public benchmark datasets in Table~\ref{table_bench}. 
	The evaluation of the self-ensemble is also provided in the last two columns. We trained our models using $10^{6}$ updates with batch size 16. We keep the other settings same as the baseline models. 
	We compare our models with the state-of-the-art methods including A+~\cite{timofte2014a+}, SRCNN~\cite{dong2014learning}, VDSR~\cite{kim2016accurate}, and SRResNet~\cite{ledig2016photo}. For comparison, we measure PSNR and SSIM on the y channel and ignore the same amount of pixels as scales from the border.
	We used MATLAB \cite{MATLAB:2016} functions for evaluation. Comparative results on DVI2K dataset are also provided. Our models exhibit a significant improvement compared to the other methods. The gaps further increase after performing self-ensemble. We also present the qualitative results in Fig.~\ref{fig_result_1}. The proposed models successfully reconstruct the detailed textures and edges in the HR images and exhibit better-looking SR outputs compared with the previous works. %We provide the full test images in the supplementary material.     

	\section{NTIRE2017 SR Challenge}
	
	This work is initially proposed for the purpose of participating in the NTIRE2017 Super-Resolution Challenge ~\cite{Timofte_2017_CVPR_Workshops}.
	The challenge aims to develop a single image super-resolution system with the highest PSNR. 
	% 	While dealing with the traditional bicubic downsampling method, the challenge also suggest the needs for more robust super-resolution algorithms with unknown downsampling track.
	
\begin{figure*}[t]
		\captionsetup[subfloat]{labelformat=empty}
		\begin{center}
			\newcommand{\rowArg}{2.46cm}
			\newcommand{\fullSize}{5.56cm}
			\newcommand{\fullHalf}{2.78cm}
			\newcommand{\patchSize}{2.3cm}
			\scriptsize
			\setlength\tabcolsep{0.1cm}
			\begin{tabular}[b]{c c}
				\begin{tabular}[b]{c c c}
					\multirow{2}{*}[\rowArg]{
						\subfloat[0791 from DIV2K~\cite{Timofte_2017_CVPR_Workshops}]
						{\includegraphics[width = \fullHalf, height = \fullSize]
							{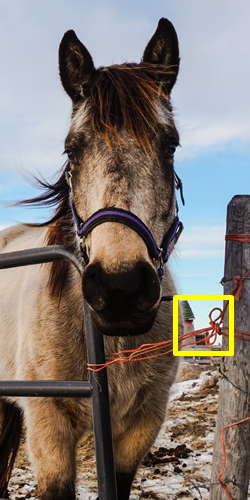}}} &
					\subfloat[HR \protect\linebreak(PSNR / SSIM)]
					{\includegraphics[width = \patchSize, height = \patchSize]
						{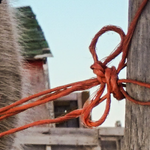}} &
					\subfloat[Bicubic \protect\linebreak(22.20 dB / 0.7979)]
					{\includegraphics[width = \patchSize, height = \patchSize]
						{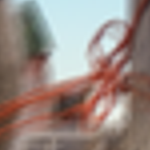}} \\ [-0.2cm] &
					\subfloat[\textbf{EDSR (Ours)} \protect\linebreak(\textcolor{red}{29.05 dB} / \textcolor{red}{0.9257})]
					{\includegraphics[width = \patchSize, height = \patchSize]
						{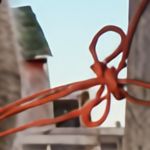}} &
					\subfloat[\textbf{MDSR (Ours)} \protect\linebreak(\textcolor{blue}{28.96 dB} / \textcolor{blue}{0.9244})]
					{\includegraphics[width = \patchSize, height = \patchSize]
						{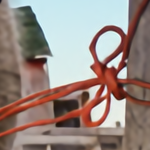}} 
				\end{tabular}
				&  
				\begin{tabular}[b]{c c c}
					\multirow{2}{*}[\rowArg]{
						\subfloat[0792 from DIV2K~\cite{Timofte_2017_CVPR_Workshops}]
						{\includegraphics[width = \fullHalf, height = \fullSize]
							{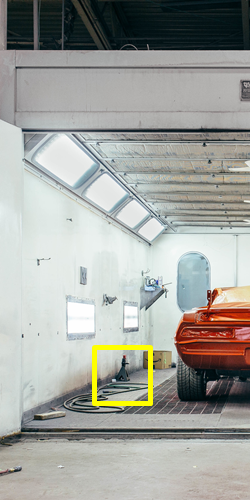}}} &
					\subfloat[HR \protect\linebreak(PSNR / SSIM)]
					{\includegraphics[width = \patchSize, height = \patchSize]
						{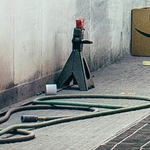}} &
					\subfloat[Bicubic \protect\linebreak(21.59 dB / 0.6846)]
					{\includegraphics[width = \patchSize, height = \patchSize]
						{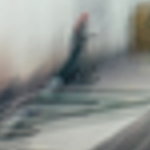}} \\ [-0.2cm] &
					\subfloat[\textbf{EDSR (Ours)} \protect\linebreak(\textcolor{red}{27.24 dB} / \textcolor{red}{0.8376})]
					{\includegraphics[width = \patchSize, height = \patchSize]
						{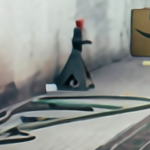}} &
					\subfloat[\textbf{MDSR (Ours)} \protect\linebreak(\textcolor{blue}{27.14 dB} / \textcolor{blue}{0.8356})]
					{\includegraphics[width = \patchSize, height = \patchSize]
						{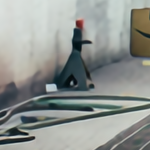}} 
				\end{tabular}
			\end{tabular}
			
			\begin{tabular}[b]{c c}
				\begin{tabular}[b]{c c c}
					\multirow{2}{*}[\rowArg]{
						\subfloat[0793 from DIV2K~\cite{Timofte_2017_CVPR_Workshops}]
						{\includegraphics[width = \fullHalf, height = \fullSize]
							{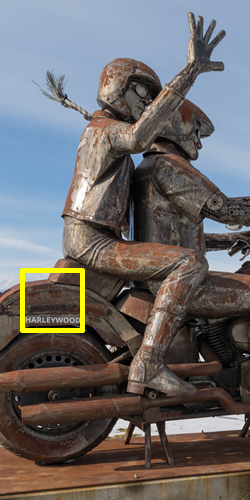}}} &
					\subfloat[HR \protect\linebreak(PSNR / SSIM)]
					{\includegraphics[width = \patchSize, height = \patchSize]
						{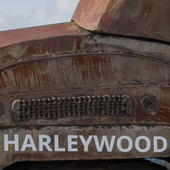}} &
					\subfloat[Bicubic \protect\linebreak(23.81 dB / 0.8053)]
					{\includegraphics[width = \patchSize, height = \patchSize]
						{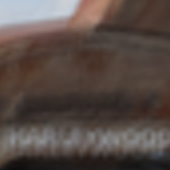}} \\ [-0.2cm] &
					\subfloat[\textbf{EDSR (Ours)} \protect\linebreak(\textcolor{red}{30.94 dB} / \textcolor{red}{0.9318})]
					{\includegraphics[width = \patchSize, height = \patchSize]
						{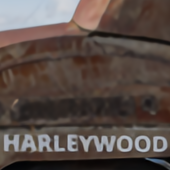}} &
					\subfloat[\textbf{MDSR (Ours)} \protect\linebreak(\textcolor{blue}{30.81 dB} / \textcolor{blue}{0.9301})]
					{\includegraphics[width = \patchSize, height = \patchSize]
						{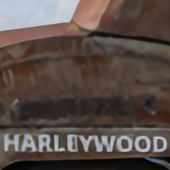}} \\
				\end{tabular}
				&  
				\begin{tabular}[b]{c c c}
					\multirow{2}{*}[\rowArg]{
						\subfloat[0797 from DIV2K~\cite{Timofte_2017_CVPR_Workshops}]
						{\includegraphics[width = \fullHalf, height = \fullSize]
							{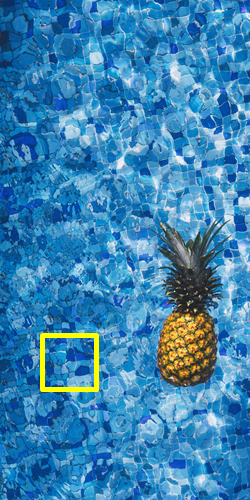}}} &
					\subfloat[HR \protect\linebreak(PSNR / SSIM)]
					{\includegraphics[width = \patchSize, height = \patchSize]
						{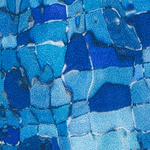}} &
					\subfloat[Bicubic \protect\linebreak(19.77 dB / 0.8937)]
					{\includegraphics[width = \patchSize, height = \patchSize]
						{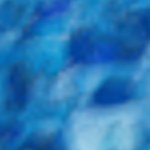}} \\ [-0.2cm] &
					\subfloat[\textbf{EDSR (Ours)} \protect\linebreak(\textcolor{red}{25.48 dB} / \textcolor{red}{0.9597})]
					{\includegraphics[width = \patchSize, height = \patchSize]
						{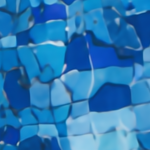}} &
					\subfloat[\textbf{MDSR (Ours)} \protect\linebreak(\textcolor{blue}{25.38 dB} / \textcolor{blue}{0.9590})]
					{\includegraphics[width = \patchSize, height = \patchSize]
						{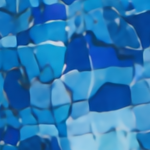}} \\
				\end{tabular}
			\end{tabular}
		\end{center}
		\captionsetup{justification=raggedright,singlelinecheck=false}
		\caption{Our NTIRE2017 Super-Resolution Challenge results on unknown downscaling $\times 4$ category. In the challenge, we excluded images from 0791 to 0800 from training for validation. We did not use geometric self-ensemble for unknown downscaling category.}
		\label{fig_result_unknown}
	\end{figure*}

	In the challenge, there exist two tracks for different degraders (bicubic, unknown) with three downsample scales $(\times 2, 3, 4)$ each.
	Input images for the unknown track are not only downscaled but also suffer from severe blurring.
	Therefore, more robust mechanisms are required to deal with the second track.
	We submitted our two SR models (EDSR and MDSR) for each competition and prove that our algorithms are very robust to different downsampling conditions.
	Some results of our algorithms on the unknown downsampling track are illustrated in Fig.~\ref{fig_result_unknown}. Our methods successfully reconstruct high-resolution images from severely degraded input images.
	Our proposed EDSR+ and MDSR+ won the first and second places, respectively, with outstanding performances as shown in Table~\ref{table_ntire_result}. 
	
	\begin{table*}[h]
		{\footnotesize
			\setlength\tabcolsep{8pt}
			\renewcommand{\arraystretch}{1.3}
			\begin{center}
				\begin{tabular}{c*{6}{c}*{6}{c}}
					\hline
					& \multicolumn{6}{c}{Track1: bicubic downscailing} 
					& \multicolumn{6}{c}{Track2: unknown downscailing} \\
					& \multicolumn{2}{c}{$\times 2$}
					& \multicolumn{2}{c}{$\times 3$} 
					& \multicolumn{2}{c}{$\times 4$} 
					& \multicolumn{2}{c}{$\times 2$} 
					& \multicolumn{2}{c}{$\times 3$} 
					& \multicolumn{2}{c}{$\times 4$} \\
					Method &
					PSNR & SSIM & PSNR & SSIM & PSNR & SSIM & PSNR & SSIM &	PSNR & SSIM & PSNR & SSIM \\
					\hline
					\textbf{EDSR+} (Ours) &
					\textcolor{red}{34.93} & \textcolor{red}{0.948} &
					\textcolor{red}{31.13} & \textcolor{red}{0.889} &
					\textcolor{red}{29.09} & \textcolor{red}{0.837} &
					\textcolor{red}{34.00} & \textcolor{red}{0.934} &
					\textcolor{red}{30.78} & \textcolor{red}{0.881} &
					\textcolor{red}{28.77} & \textcolor{red}{0.826} \\
					\textbf{MDSR+} (Ours) &
					\textcolor{blue}{34.83} & \textcolor{blue}{0.947} &
					\textcolor{blue}{31.04} & \textcolor{blue}{0.888} &
					\textcolor{blue}{29.04} & \textcolor{blue}{0.836} &
					\textcolor{blue}{33.86} & \textcolor{blue}{0.932} &
					\textcolor{blue}{30.67} & \textcolor{blue}{0.879} &
					\textcolor{blue}{28.62} & \textcolor{blue}{0.821} \\
					3rd method &
					{34.47} & {0.944} &
					{30.77} & {0.882} &
					{28.82} & {0.830} &
					{33.67} & {0.930} &
					{30.51} & {0.876} &
					{28.54} & {0.819} \\
					4th method &
					{34.66} & {0.946} &
					{30.83} & {0.884} &
					{28.83} & {0.830} &
					{32.92} & {0.921} &
					{30.31} & {0.871} &
					{28.14} & {0.807} \\
					5th method &
					{34.29} & {0.948} &
					{30.52} & {0.889} &
					{28.55} & {0.752} &
					{-} & {-} &
					{-} & {-} &
					{-} & {-} \\
					\hline
				\end{tabular}
			\end{center}
		}
		\captionsetup{justification=raggedright,singlelinecheck=false}
		\caption{Performance of our methods on the test dataset of NTIRE2017 Super-Resolution Challenge  \cite{Timofte_2017_CVPR_Workshops}. The results of top 5 methods are displayed for two tracks and six categories. Red indicates the best performance and blue indicates the second best. 
		}
		\label{table_ntire_result}
	\end{table*}
	
	\section{Conclusion}
	
	In this paper, we proposed an enhanced super-resolution algorithm. By removing unnecessary modules from conventional ResNet architecture, we achieve improved results while making our model compact. We also employ residual scaling techniques to stably train large models. Our proposed singe-scale model surpasses current models and achieves the state-of-the-art performance.
	
	Furthermore, we develop a multi-scale super-resolution network to reduce the model size and training time.
	With scale-dependent modules and shared main network, our multi-scale model can effectively deal with various scales of super-resolution in a unified framework. While the multi-scale model remains compact compared with a set of single-scale models, it shows comparable performance to the single-scale SR model.%, only with 6-times fewer parameters.
	
	Our proposed single-scale and multi-scale models have achieved the top ranks in both the standard benchmark datasets and the DIV2K dataset.
	
	%\clearpage
	{\small
		\bibliographystyle{ieee}
		\bibliography{egbib}
	}
	
\end{document}